\documentclass[runningheads]{llncs}
\usepackage[T1]{fontenc}
\usepackage{graphicx}
\usepackage{booktabs}
\usepackage{amsmath}
\usepackage{xcolor}
\usepackage{soul}
\begin{document}
\title{Analysis of  Ethnic Disparities in Autism Spectrum Disorder among Toddlers}
\author{Aadithya Prabha Ramaharsha  \and Deevna Reddy \and Uma Ranjan \orcidID{0000-0001-6258-4513}}
\institute{Sri Ramachandra Institute of Higher Education and Research}
\maketitle
\begin{abstract}
Autism Spectrum Disorder (ASD) is a neurodevelopmental disorder characterized by challenges in communication and behavior. This study examines the relationship between ethnicity and ASD traits, along with  behavioural scores, sex and neonatal jaundice across three ethnic groups: White Europeans, Asians, and Middle Eastern individuals. We perform a logistic regression and show that ethnicity has a signifcant effect on incidence of ASD. White Europeans are 81\% increased risk of ASD
and Middle Easterners are at 79\% reduced risk of ASD compared to Asians. 
We also confirm earlier studied which show that neonatal jaundice is a significant predictor of ASD, while male children are at much higher risk of ASD compared to female children. 

These results suggest the need for diagnostic frameworks and interventions that account for ethnic in the presentation and assessment of ASD traits. 

\end{abstract}
\section{Introduction}
Autism Spectrum Disorder (ASD) is a complex neurodevelopmental condition that profoundly impacts communication, social interaction, and behavior. Often characterized by repetitive actions and restricted interests, ASD typically manifests in early childhood, with symptoms emerging within the first few years of life. Early diagnosis and timely intervention are critical for improving developmental outcomes, yet diagnosing ASD remains a challenge due to its diverse presentation and the influence of genetic, environmental, and cultural factors.

Autism manifests across a spectrum, with both congenital and acquired mechanisms. 
Congenital autism typically manifests from birth and is associated with genetic and prenatal factors, such as mutations in genes like CHD8 and maternal health conditions \cite{plandrigan,ssandin}. Acquired autism, on the other hand, emerges later in childhood, often influenced by postnatal environmental factors, including parental age, exposure to pollutants, and early social experiences. However, even in the case of acquired autism, genetic make-up plays a significant role. Hence, the effect of ethnicity as a causative reason for autism is important to study.

Autism is diagnosed primarily through behavioural traits. These traits are captured in the form of a questionnaire (A1 to A10) based on the Q-CHAT-10 (Quantitative Checklist for Autism in Toddlers – 10 items), a widely used and validated screening tool developed by researchers at the Autism Research Centre, University of Cambridge \cite{callison}. The Q-CHAT-10 is designed to detect early behavioral signs of autism in toddlers by asking caregivers a series of structured questions. Each of the items (A1 to A10) corresponds to specific developmental and social behaviors such as eye contact, response to name, imitation, use of gestures, and interest in social interaction. These responses are typically scored to assess the presence and severity of autistic traits.

In addition to behavioural traits, demograohic factors like sex and health related factors, such neonatal jaundice have also been shown to be linked to autism. For instance, Boys with neonatal jaundice are at highest risk of developing autism. In addition, prevalence studies of autism have suggested that variations across Ethnicities.  These differences are partially attributed to cultural differences in perceiving autism symptoms, along with uneven access to healthcare, which delay diagnosis for many children. Addressing these challenges requires diagnostic frameworks that account for ethnic and cultural diversity, ensuring equitable identification and support for all individuals with ASD.  It was initially believed that lower rates of incidence among non-white population could arise from lower levels of screening. However, Recent data from the CDC reveal that identification rates among Black, Hispanic, and Asian or Pacific Islander children now surpass those of White children, reflecting strides in awareness and access to services in these communities. Despite improvements in awareness, disparities among different groups in prevalence of autism persist. 

Understanding how ethnicity influences ASD traits is crucial for developing effective, inclusive diagnostic tools and interventions. By recognizing and addressing cultural differences, outcomes can be improved for individuals with ASD across diverse populations with a view  to achieving equitable care for all.




\section{Related Work}
Autism Spectrum Disorder (ASD) research has been studied extensively, with increasing focus on leveraging machine learning (ML) for diagnosis, and addressing ethnic disparities in ASD diagnosis and care. This section synthesizes existing work across these domains, identifies critical gaps, and positions the current study within this broader context.

Machine learning has emerged as a powerful tool for enhancing ASD diagnostics, offering the ability to analyze complex behavioral, genetic, and neuroimaging data to improve early detection and classification. Farooq et al. \cite{farooq2023} highlighted the integration of ML in diagnosing ASD across children and adults, emphasizing its role in standardizing assessments using diverse datasets. Similarly, Vakadkar et al. \cite{kvakadhar} demonstrated how feature extraction from clinical datasets improved predictive accuracy for ASD traits in children, underscoring ML’s potential in addressing diagnostic variability. Sharma et al.\cite{psharma} reviewed ML algorithms applied in autism diagnosis, providing insights into supervised, unsupervised, and hybrid approaches, and their comparative effectiveness. Zhang et al.\cite{yzhang} conducted a systematic review of ML-based ASD prediction, identifying significant gaps in the generalizability of models across ethnically diverse populations.

Recent advancements in behavioral data modeling have also shown promise. Alharthy et al.\cite{alharthy} proposed a novel ML framework using behavioral datasets, achieving high accuracy in predicting ASD traits. Mohammed et al.\cite{amohammed} and Hussain et al.\cite{mhussain} offered comprehensive overviews of ML-based diagnostic tools, focusing on the integration of multimodal data, such as genetic predispositions and environmental triggers. Despite these advances, challenges persist, particularly in interpreting models and adapting them to diverse cultural contexts.

Ye et al.\cite{toxicity} identified environmental toxins as a significant contributor to the onset of acquired autism, while Sandin et al. \cite{ssandin} explored how parental age correlates with increased risk. Acquired autism is particularly challenging to diagnose, as its later onset can lead to delays in detection and intervention, exacerbating developmental impacts. 

 Mandell et al.\cite{dsmandell} reported persistent disparities in diagnosis rates, with Black and Hispanic children experiencing significant delays compared to their White counterparts. Recent data from the CDC indicates increasing identification rates among minority groups, reflecting improved awareness but underscoring the need for equitable access to diagnostic and intervention services.

Cultural perceptions of autism symptoms vary significantly. For example, behaviors perceived as atypical in one cultural context may be normalized in another, leading to underdiagnosis among certain ethnic groups (Kim et al.\cite{ykim}). Diagnostic tools like ADOS-2 and M-CHAT, while widely used, have faced criticism for cultural and linguistic biases (Smith et al.\cite{tsmith}). Studies emphasize the importance of adapting these tools to better capture culturally nuanced expressions of ASD traits. Systemic barriers, such as socioeconomic inequities and language differences, further compound ethnic disparities, limiting early detection and intervention opportunities (Zuckerman et al.\cite{kzuckerman}).

Despite substantial progress, notable gaps remain. Most ML-based studies rely on homogeneous datasets, limiting their applicability to diverse populations. 

This study looks at how ethnicity might influence ASD traits, focusing on three groups: White European, South Asian, and Middle Eastern populations. Through regression analysis, it explores the relationship between ethnicity, behavioral traits, and Qchat scores to better understand cultural differences in how ASD may present. The goal is to help shape diagnostic practices that take these differences into account, making them more inclusive and accessible to people from all backgrounds.

\section{Proposed Method}
\subsection{Data Sources}
The data for this study was taken from the Toddler Autism Dataset (July 2018), comprising 3648 instances with a variety of features, including demographic details (age, sex, ethnicity), behavioral assessments (Qchat-10 scores), and clinical diagnostic results for toddlers aged 12-36 months. This dataset also captures health-related variables, such as jaundice status and family history of ASD.

\subsection{Data Curation}

The primary dataset used in this study, the Toddler Autism Dataset, consists of the following key variables:

\begin{itemize}
    \item \textbf{Q-CHAT-10 Score}: An autism screening questionnaire score ranging from 0-10, with higher scores indicating a higher likelihood of ASD traits. This is the primary outcome variable.
    \item \textbf{Age}: The age of each toddler in months at the time of screening, ranging from 12-36 months.
    \item \textbf{Sex}: The biological sex of each toddler, with 55\% male and 45\% female.
    \item \textbf{Ethnicity}: The ethnicity of each toddler, including White European, Middle Eastern, Asian, Black, Hispanic, Native Indian, Latino, South Asian, and Mixed.
    \item \textbf{History of Jaundice}: Whether the toddler had a history of jaundice at birth (yes/no).
    \item \textbf{Family History of ASD}: Whether a family member of the child had been diagnosed with ASD (yes/no).
    \item \textbf{Test Administrator}: Whether the screening test was completed by a family member or a healthcare professional.
\end{itemize}

The ASD diagnosis was done through a separate clinical procedure. 

\subsection{Data Preprocessing}

  An initial exploratory analysis of the data was performed to check correlations and distributions. Based on this, the ethnicities which had low frequencies or did not correspond to a single ethnicity (such as 'Others' or 'Mixed') were dropped. Three values of ethnicity qualified for further analysis — White European, Middle Eastern, and Asian.
  Further, the QChat score which was a sum of the  scores A1 - A10 was also excluded, since the individual scores were considered. The categorical variables such as jaundice, sex and ethnicity were represented through one-hot encoding. The values of A1 - A10 were considered individual variables, representing 1 or 0.
  
\subsection{Exploratory Data Analysis (EDA)}
Exploratory data analysis was conducted using box plots to visualize the distribution of Qchat scores across different ethnicities. The mean scores were calculated for each ethnic group to identify potential disparities.

\subsection{Statistical Analysis}

Statistical tests to check both unadjusted and adjusted correlations of ethnicity with autism were performed. 

Chi-squared tests were conducted to explore  unadjusted associations between categorical variables such as sex, ethnicity, and jaundice status with ASD traits, according to  Equation~\ref{eqn:chi_square},

\begin{equation}
    \chi^2 = \sum [\frac{(O-E)^2}{E}]
    \label{eqn:chi_square}
\end{equation}

where \( O \) represents observed frequencies and \( E \) represents expected frequencies. Since the p-value is less than 0.05, we conclude that there is a statistically significant association between sex and ASD traits.

Logistic Regression was performed to check  autism from the chosen factors, after adjustment for other factors. The logit transformation was applied to the selected features 

\[
    Logit(p) = \log\left(\frac{p}{1-p}\right)
\]

where \( p \) represents the probability of success. The transformed variables were then analyzed to assess their distributions.

For Sex, Female was chosen as the reference variable and for ethnicity, Asian was chosen as reference variable. Bonferroni correction was applied to the p-values to check significance.

Since autism traits are detected primarilty through the mean of the QChat scores, a Analysis of Variance (ANOVA) to investigate significant differences of the QChat mean score across categories.

\subsection{Variance Inflation Factor (VIF) Assessment}
To assess multicollinearity among features in our models, we calculated Variance Inflation Factor (VIF) for each variable.

\begin{equation}
    VIF = \frac{1}{1-R^2}
\end{equation}

A VIF value greater than 5 indicates a high degree of multicollinearity among predictors.

\begin{figure}[ht]
    \centering
    \includegraphics[width=0.6\linewidth]{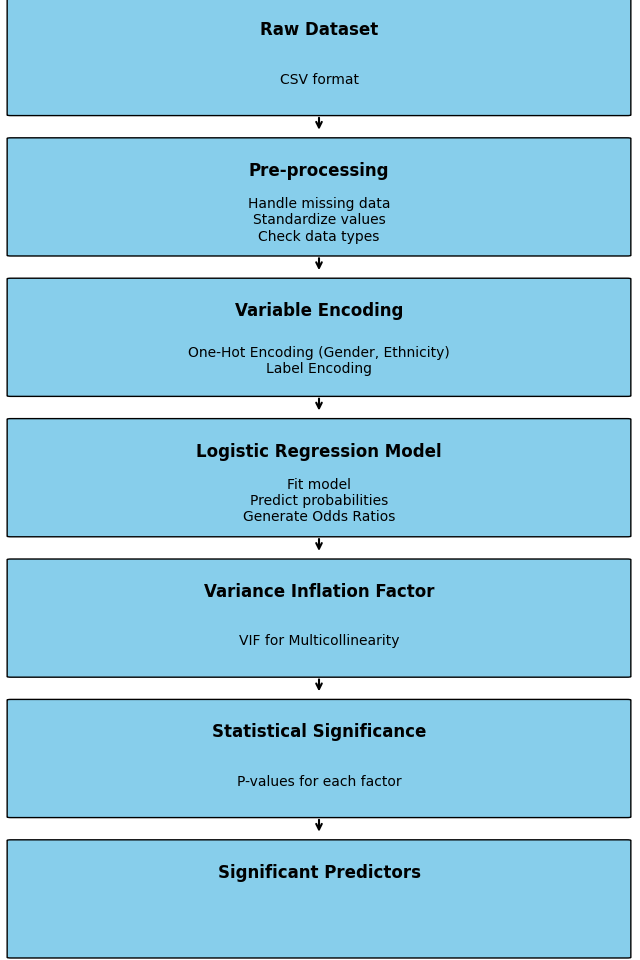}
    \caption{Analysis Procedure}
    \label{fig:Workflow}
\end{figure}

The results of logistic regression were compared with those of the statistical tests, to check for the detection of autism with and without adjusted factors. The overall procedure is shown in  Figure~\ref{fig:Workflow}

\section{Results}

The results include initial descriptive statistics, as well as statistical results. 

\subsection{Descriptive Statistics}
\begin{figure}[htbp]
    \centering
    \includegraphics[width=0.5\linewidth]{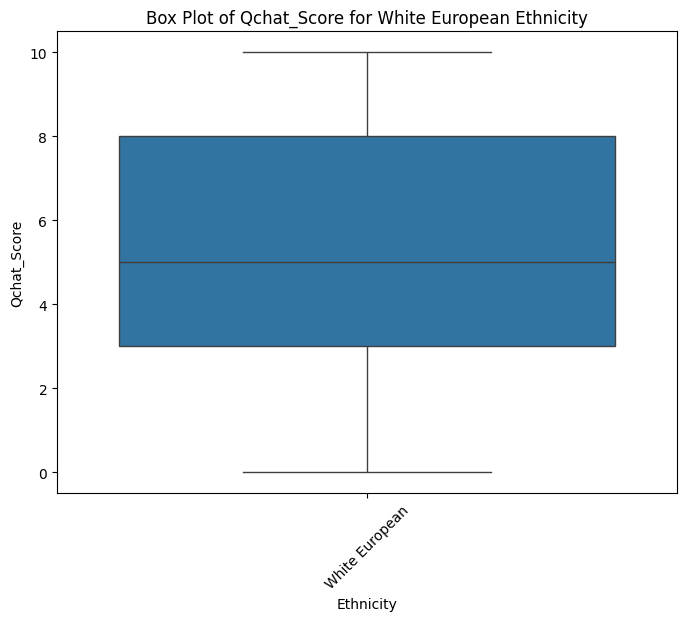}
    \caption{Qchat score for White European Ethnicity}
    \label{fig:boxplot_white_european}
\end{figure}

\begin{figure}[htbp]
    \centering
    \includegraphics[width=0.5\linewidth]{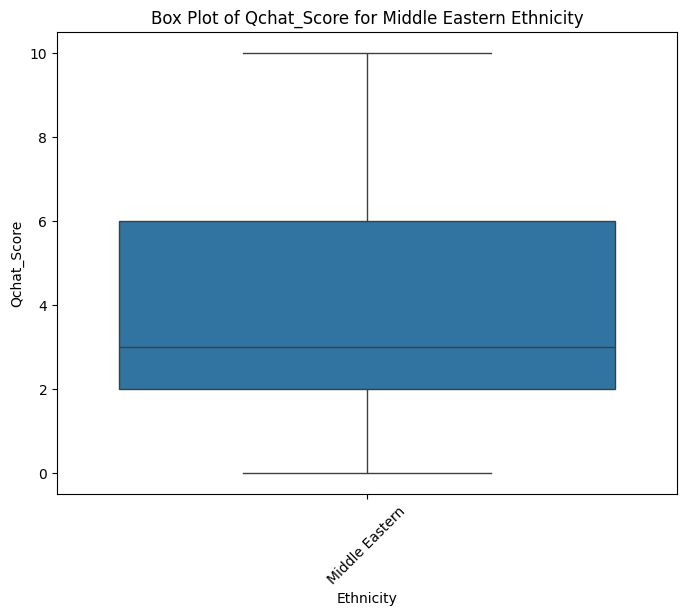}
    \caption{Qchat score for Middle Eastern Ethnicity}
    \label{fig:boxplot_middle_eastern}
\end{figure}

\begin{figure}[htbp]
    \centering
    \includegraphics[width=0.5\linewidth]{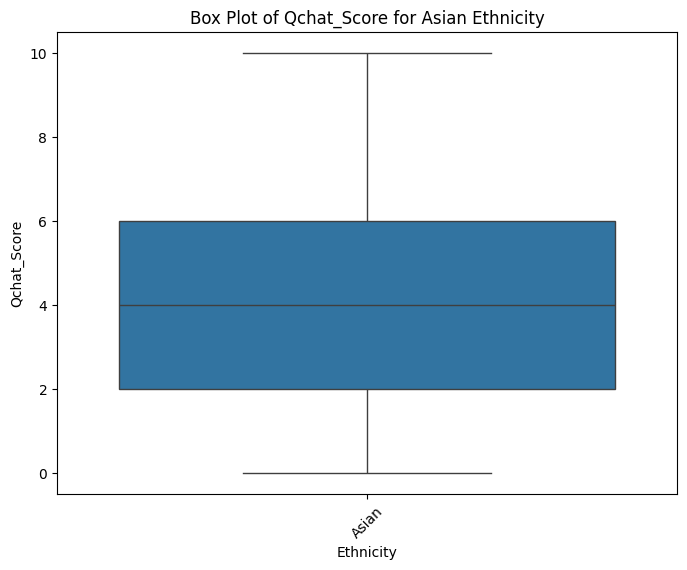}
    \caption{Qchat score for Asian Ethnicity}
    \label{fig:boxplot_asian}
\end{figure}

The dataset included a total of 3648 instances of toddlers. Of these, the distribution of  Qchat scores across the various ethnicities are summarized in Figures~\ref{fig:boxplot_white_european},\ref{fig:boxplot_middle_eastern} and \ref{fig:boxplot_asian}. The summary statistics are presented in Table~\ref{tab:descriptive_statistics}.


\begin{table}[htbp]
    \centering
    \caption{Descriptive Statistics of Qchat Scores by Ethnicity}
    \begin{tabular}{@{}lccc@{}}
        \toprule
        Ethnicity         & Mean & Median & Std. Deviation\\ 
        \midrule
        Asian    & 4.112 & 4.0  & 2.693 \\
        Middle Eastern & 3.979  & 3.0 & 2.720       \\
        White European & 5.236 & 5.0 & 3.009         \\ 
        \bottomrule
    \end{tabular}
    \label{tab:descriptive_statistics}
\end{table}

The mean Qchat scores indicate that Asian and Milddle Eastern children have similar average score, while White European children have a higher score, suggesting a greater likelihood of ASD traits.  


\subsection{ANOVA Results}
An Analysis of Variance (ANOVA) was performed to assess the significance of differences in Qchat scores among the ethnic groups. The results indicated a statistically significant effect of ethnicity on Qchat scores, with an F-statistic of 1723.23 and a p-value of 0.0000. This result suggests that at least one ethnic group differs significantly from the others in terms of Qchat scores.

\begin{equation}
    F = \frac{\text{Between-group variance}}{\text{Within-group variance}}
\end{equation}

Given that the p-value is less than the significance level of 0.05, we reject the null hypothesis, concluding that there are significant differences in Qchat scores among the ethnic groups.


\subsection{Chi-Squared Results}

The Chi-Squared results for the univariate association are listed in 
Table~\ref{tab:chi_square_ethnicity}. It can be seen that these are not statistically significant. However, the chi-squared values are not adjusted for other factors, and hence indicate that ethnicity is not an independent predictor of ASD, and needs to be considered in conjunction with other factors. 

\begin{table}[htbp]
    \centering
    \caption{Chi-Square Test Results for Ethnicity and ASD Traits}
    \begin{tabular}{@{}lccc@{}}
        \toprule
        Ethnicity & Chi\textsuperscript{2} & p-value & Degrees of Freedom \\
        \midrule
        Asian & 1.092301 & 0.295962 & 1 \\
        White European & 0.769037 & 0.380515 & 1 \\
        Middle Eastern & 0.016617 & 0.897432 & 1 \\
        \bottomrule
    \end{tabular}
    \label{tab:chi_square_ethnicity}
\end{table}

\subsection{Logistic Regression Results}

Results of the  logistic regression are presented in Table~\ref{tab:logisticResults} and \ref{tab:logisticOdds}. It is seen  from the coefficients that White European children are at a much higher risk and Middle Eastern children are at a much lower risk of autism compared to Asian Children, with other factors adjusted. It is seen from the p-values that both these comparisons are statistically significant. Among demographic factors, sex has a significant role to play. Some behavioural traits, notably A9 and A6,  are more important than others in predicting autism. Some traits, such as A1, A3, A8 and A10 , as deemed from the p-values after applying Bonferroni correction, are not significant. This may be an important factor to consider, since the current overall score takes a total of all the Qchat scores, assigning them equal importance.


The Variance Inflation Factors  are summarized in Table~\ref{tab:vif_results}. It can be seen that the multicollinearity effects are minimal, and hence, the coefficients truly reflect the dependence on ASD.
\begin{table}[h!]
\centering
\caption{Logit Regression Summary}
\begin{tabular}{|l|c|c|c|c|c|c|}
\hline
\textbf{Feature} & \textbf{Coef} & \textbf{Std Err} & \textbf{z} & \textbf{P$>$|z|} & \textbf{[0.025} & \textbf{0.975]} \\
        \hline
        const & -2.6089 & 0.165 & -15.802 & 0.000 & -2.933 & -2.285 \\
        A1 & -0.3071 & 0.141 & -2.171 & 0.030 & -0.584 & -0.030 \\
        A2 & 0.6327 & 0.148 & 4.265 & 0.000 & 0.342 & 0.923 \\
        A3 & -0.0464 & 0.163 & -0.284 & 0.776 & -0.366 & 0.273 \\
        A4 & 0.3410 & 0.154 & 2.211 & 0.027 & 0.039 & 0.643 \\
        A5 & 0.8642 & 0.154 & 5.625 & 0.000 & 0.563 & 1.165 \\
        A6 & 1.4911 & 0.149 & 9.983 & 0.000 & 1.198 & 1.784 \\
        A7 & 0.7362 & 0.138 & 5.344 & 0.000 & 0.466 & 1.006 \\
        A8 & -0.0251 & 0.137 & -0.184 & 0.854 & -0.293 & 0.243 \\
        A9 & 1.6730 & 0.129 & 12.986 & 0.000 & 1.342 & 2.004 \\
        A10 & 0.0638 & 0.120 & 0.532 & 0.595 & -0.171 & 0.299 \\
        sex\_male & 1.5826 & 0.133 & 11.881 & 0.000 & 1.322 & 1.844 \\
        jaundice\_yes & 0.5755 & 0.111 & 5.199 & 0.000 & 0.359 & 0.792 \\
        ethnicity\_White European & 0.5961 & 0.122 & 4.867 & 0.000 & 0.356 & 0.836 \\
        ethnicity\_Middle Eastern & -1.5452 & 0.150 & -10.331 & 0.000 & -1.838 & -1.252 \\
         \hline
\end{tabular}
\label{tab:logisticResults}
\end{table}

\begin{table}[h!]
\centering
\caption{Odds Ratios of Predictive Features}
\begin{tabular}{|l|c|c|}
\hline
\textbf{Feature} & \textbf{Odds Ratio} & \textbf{P-Value} \\
\hline
A9 & 5.328356 & 3.468618e-23 \\
sex\_male & 4.867571 & 1.481147e-32 \\
A6 & 4.441848 & 1.803146e-23 \\
A5 & 2.373195 & 1.850896e-08 \\
A7 & 2.087899 & 9.088598e-08 \\
A2 & 1.882742 & 1.998366e-05 \\
ethnicity\_White European & 1.815113 & 1.132929e-06 \\
jaundice\_yes & 1.789856 & 2.005046e-07 \\
A4 & 1.406330 & 2.703489e-02 \\
A10 & 1.065895 & 5.945189e-01 \\
A8 & 0.975188 & 8.542383e-01 \\
A3 & 0.954661 & 7.761013e-01 \\
A1 & 0.735610 & 2.618086e-02 \\
ethnicity\_Middle Eastern & 0.573525 & 5.117813e-25 \\
const & 0.073612 & 3.010626e-56 \\
\hline
\end{tabular}
\label{tab:logisticOdds}
\end{table}

\begin{table}[htbp]
    \centering
    \caption{Variance Inflation Factor (VIF) Results}
    \begin{tabular}{@{}lc@{}}
        \toprule
        Feature & VIF \\ 
        \midrule
        A1 & 1.717832 \\ 
        A2 & 1.514004 \\ 
        A3 & 1.722339 \\ 
        A4 & 1.964504 \\ 
        A5 & 1.922325 \\ 
        A6 & 1.939300 \\ 
        A7 & 1.776987 \\ 
        A8 & 1.390746 \\ 
        A9 & 1.989263 \\ 
        A10 & 1.217635 \\ 
        Male & 1.091710 \\ 
        Jaundice & 1.062811 \\ 
        White European & 1.357859 \\ 
        Middle Eastern & 1.377278 \\
        \bottomrule
    \end{tabular}
    \label{tab:vif_results}
\end{table}

\subsection{Summary}

The results indicate that ethnicity has a significant association with ASD diagnosis in a multivariate setting, despite failing to achieve significance in univariate chi-square tests. This highlights the importance of considering multivariate models, particularly when potential confounders or interactions may mask true effects.

The logistic regression model revealed that individuals identified as White European 

Other features, such as gender (male), specific behavioral indicators (A9, A6, A5), and history of jaundice, also emerged as strong predictors. The elevated odds for male participants are consistent with existing literature on ASD prevalence. Similarly, behavioral traits (A1–A10) showed varying degrees of association, with A9 and A6 having the strongest influence, both with high odds ratios and very low p-values.

The divergence in findings between chi-square and logistic regression underscores the limitations of univariate analyses for complex conditions such as ASD. Chi-square tests ignore the potential interaction and confounding effects that are central to multivariable modeling. Our results demonstrate how ethnic differences become statistically relevant only when adjusted for co-predictors like behavioral indicators and gender.

Finally, the low VIF values across all predictors reinforce the reliability of the model, suggesting that the estimated effects are not distorted by multicollinearity.

\section{Conclusion}

This study examines ethnic differences in Autism Spectrum Disorder (ASD) traits, highlighting the influence of cultural and demographic factors on their presentation and assessment. We show that ethnicity plays an important role in autism, and must be taken into account along with other factors such as sex, health  and behavioural traits. It is seen that compared to Asians, White Europeans are at higher risk of autism, while Middle Easterners are at lower risk of autism. The  data compiled for this study was recent, indicating that the screening rates of the various population are nearly the same. This is also reflected in the nearly equal numbers of the ethnicities considered for this study.

A secondary conclusion is in the heterogeneity of behavioural traits. While the QChat questionnaire has been used to screen for autism, and a total of the scores used to indicate autism, our analysis shows that a more nuanced interpretation may be needed, with some traits being more important than others.

\end{document}